# Cross-Asset Knowledge Transfer for Early Fault Detection of Data-Scarce Assets Without Target Fault Data

Zekai Wang[1,*], Yuxuan Li[2,*], Zhangyue Shi[3,+]

[1] Charles F. Dolan School of Business, Fairfield University, Fairfield, CT, United States

[2] School of Economics and Management, Tiangong University, Tianjin, China

[3] Independent Researcher, Chesterfield, MO, United States

* Zekai Wang and Yuxuan Li contributed equally to this work.

[+] Corresponding author: zhangyue.shi@outlook.com

**Abstract**

Reliability management of industrial fleets depends on early fault warnings, yet lifecycle data across similar assets are highly imbalanced: a few assets accumulate run-to-failure histories, while newly installed or rarely failing assets provide only normal-condition data. This paper asks how diagnostic knowledge can be transferred from a data-rich source asset to a data-scarce target asset when no fault data have been observed on the target. We formulate cross-asset early fault detection as a zero-shot domain adaptation problem, in which zero-shot refers to the absence of target fault samples, and propose a framework that requires only normal data from the target. The framework comprises (1) a statistical process control step that labels the source asset's run-to-failure data with an exponentially weighted moving average chart, reducing manual annotation to verifying flagged segments; (2) an Adapted Domain-Adversarial Neural Network (ADANN) that combines domain-adversarial training and a Maximum Mean Discrepancy penalty with an Adapted Classification-Domain (ACD) loss, which anchors the classifier's normal region to the target's own normal data; and (3) an ahead-of-time labeling scheme with a prediction horizon that targets incipient faults. Under an exact-classification assumption, a theoretical analysis characterizes the domain-adversarial term for a normal-only target and the alignment state in which source fault features remain separated from target normal features. On a public run-to-failure bearing dataset, ADANN attains the highest mean F1-score among four transfer baselines, with a nominally significant precision gain over the strongest baseline, and its robustness is examined under synthetic noise.

## 1 Introduction

Rotating machinery is central to a wide range of industrial systems, and rolling bearings are among its most failure-prone components, accounting for a substantial portion of unplanned downtime and maintenance cost (Randall & Antoni, 2011). Condition-based maintenance addresses this risk by triggering inspection or replacement when monitored signals indicate degradation, rather than on a fixed schedule (Jardine et al., 2006; J. Lee et al., 2015). Its value depends on the detector that issues the warning: an early and reliable alarm opens a planning window for maintenance, a false alarm triggers an unnecessary intervention, and a missed or late detection exposes the system to failure. Vibration-based condition monitoring with machine learning models has therefore become a standard approach for real-time fault detection (R. Zhao et al., 2019).

In practice, however, a fundamental obstacle persists: across a fleet of similar assets, the available lifecycle data are highly imbalanced. A few assets may have complete run-to-failure histories, while many others, in particular newly installed assets or assets that are replaced preventively, operate under prolonged normal conditions with little or no recorded fault data (Lei et al., 2020). This asymmetry severely limits the diagnostic capability of models trained on the data-scarce assets. These are exactly the assets for which a maintenance decision must still be made. It raises a central question for reliability management: how can diagnostic knowledge be transferred from a data-rich source asset to a data-scarce target asset to enable early fault detection when the target has no observed fault data?

Existing research in bearing fault diagnosis has largely followed single-asset modeling paradigms, where detection models are trained and deployed independently for each bearing (Lei et al., 2020). Although transfer learning and domain adaptation techniques have been introduced to mitigate distributional shifts across operating conditions and machines (Guo et al., 2019; Yang et al., 2019), most approaches assume that the target domain contains samples of all health states, labeled or unlabeled, or at least a few labeled fault samples (Wu et al., 2020). Federated and collaborative frameworks primarily emphasize distributed or privacy-preserving learning, without explicitly addressing detection when the target asset possesses only normal operational data. Moreover, early-stage fault detection at the incipient degradation phase remains insufficiently explored under such data-scarce settings.

These limitations reveal a methodological gap: cross-domain diagnosis methods that rely on target fault samples cannot be trained when the target domain contains zero fault samples and only early-stage normal data. To address this gap, we propose a zero-shot cross-asset framework that transfers diagnostic knowledge from a data-rich source asset to a data-scarce target asset, requiring only the target's early-stage normal data. In this paper, zero-shot specifically refers to the complete absence of target-domain fault samples, rather than the absence of all target-domain data. The central difficulty is that, without any target fault examples, a model must both align the two domains' feature distributions and calibrate a fault decision boundary for a target whose faulty behavior was never observed. Our framework addresses this difficulty in three steps. It first derives fault labels on the source asset with a control chart, which reduces manual annotation to the verification of flagged segments. It then transfers the resulting diagnostic knowledge to the target through an adversarial adaptation network whose loss design exploits the limited normal target data. Finally, an ahead-of-time labeling formulation reorients the model toward anticipating incipient degradation, so that an alarm precedes the fault by a prediction horizon that can be used for maintenance planning. Together, these elements make the pipeline operable under the strict constraint that target-domain fault data are entirely unavailable.

The contributions of this work are summarized as follows:

- *A zero-shot cross-asset formulation.* We frame early fault detection on a data-scarce asset as zero-shot domain adaptation, transferring knowledge from a data-rich source asset to a target with only normal data and no fault samples. This is a stricter setting than the few-shot regimes assumed by much of the prior work on cross-domain fault diagnosis.
- *An Adapted Classification-Domain (ACD) loss with a theoretical characterization.* Within ADANN, the ACD loss uses the limited normal target data to anchor the classifier's normal region to the target, addressing a limitation of standard domain-adversarial and discrepancy-based alignment, which align features at the marginal level but leave the decision boundary fitted to the source. We characterize the equilibrium of the resulting adversarial game under a normal-only target (Theorem 1) and bound the empirical false-alarm rate on the target adaptation windows through the ACD loss, whereas no target-side control of missed detections is possible without target fault data (Remark 1).

- *An ahead-of-time detection formulation.* A sliding-window scheme with a prediction horizon reframes detection as anticipating incipient degradation, so that the model is trained to flag a future labeled fault state.

The framework is evaluated on a publicly available run-to-failure dataset with one source-target pair of deep groove ball bearings of different types operated under different radial loads, complemented by controlled noise experiments. The remainder of this paper is organized as follows. Section 2 reviews related work. Section 3 presents the methodology and its theoretical analysis. Section 4 reports the case study, including an illustrative analysis of maintenance implications. Section 5 concludes and discusses limitations and future work.

# 2 Literature Review

## 2.1 Fault Detection in Rotating Machinery

Bearing fault detection and diagnosis have long been central to predictive maintenance for rotating machinery, as bearing failures account for a substantial proportion of unplanned downtime and maintenance costs in industrial systems (Randall & Antoni, 2011). Traditional diagnostic approaches rely on signal-processing techniques such as envelope analysis, wavelet transforms, and spectral kurtosis to extract fault-sensitive features from vibration signals acquired with accelerometers (Lei et al., 2020; Randall & Antoni, 2011). While effective, these methods are typically deployed as standalone analytics on a single asset and offer no mechanism for reusing diagnostic knowledge across assets, an increasingly important capability as fleets of similar machines are instrumented and monitored at scale.

With the increasing availability of high-frequency sensor data, machine learning-based fault detection methods have gained significant attention. Supervised learning methods, including random forests, support vector machines, and deep neural networks, can achieve high diagnostic accuracy when extensive labeled datasets are available (Tang et al., 2020; R. Zhao et al., 2019). Recent studies in this journal have further developed data-driven bearing diagnosis under varying operating conditions (Udmale et al., 2026), graph-based representations of high-frequency vibration series (S. Lee et al., 2024), and robust multi-sensor fusion under missing observations (Yan et al., 2026), and some work has investigated early bearing fault detection using diffusion models (Deng et al., 2026). However, the applicability of supervised methods in real-world industrial settings is often limited, as catastrophic failures are rare and typically avoided through preventive maintenance, so that labeled fault data are scarce on most assets (Lei et al., 2020).

Unsupervised and semi-supervised approaches such as autoencoders, one-class support vector machines, isolation forests, and Gaussian mixture models learn representations of normal operating behavior and identify deviations as anomalies, which makes them attractive when labeled fault data are scarce or unavailable (J. Shi, 2023; Z. Shi et al., 2023; Tang et al., 2020; R. Zhao et al., 2019). Statistical process control (SPC) techniques, particularly univariate and multivariate control charts, have also been integrated with machine learning models to enable interpretable real-time condition monitoring (Tran et al., 2022). These hybrid frameworks leverage the statistical rigor of SPC while exploiting the representation learning capability of deep learning models, allowing for earlier detection of gradual degradation trends and process drifts.

Despite this progress, most existing studies model each asset in isolation and train a dedicated detector per machine, without any mechanism to share diagnostic knowledge across assets. This is precisely the regime that fails when a target asset has accumulated little or no fault history. How diagnostic knowledge can be transferred from a data-rich asset to a data-scarce one is the central motivation of this work and is examined in Sect. 2.2.

### 2.2 Transfer Learning and Domain Adaptation for Cross-Asset Diagnosis

When a target asset has limited fault history, a natural strategy is to transfer diagnostic knowledge from a data-rich source asset. The challenge is distribution shift: vibration data from different bearings, even of the same type, differ in distribution due to manufacturing tolerances, installation, and operating environment, so a model trained on the source degrades when applied directly to the target. Transfer learning (TL) and domain adaptation address this shift by aligning source and target representations, for example through discrepancy-based feature alignment (Gretton et al., 2006; Liu et al., 2022) and domain-adversarial training (Ganin et al., 2016). In machinery diagnostics, such methods have been used to transfer knowledge across operating conditions and machines, including from laboratory bearings to field bearings (Guo et al., 2019; Yang et al., 2019). A recurring limitation, however, is that most TL methods assume access to labeled target data, to a few target fault examples (Wu et al., 2020), or to unlabeled target data that nonetheless cover all health states. None of these is available when the target has only normal data.

Two lines of domain adaptation research are closest to the present setting. Conditional domain adaptation aligns the joint distribution of features and class predictions rather than the feature marginals alone (Long et al., 2018). In this journal, C2DAN extends this idea with entropy weighting and a supervised contrastive loss to mitigate class confusion in cross-domain anomaly detection for distributed fiber optic sensing (Xu et al., 2026). Class-conditional alignment of this kind is designed for target data that, although unlabeled, contain the classes to be aligned; with a normal-only target there is only one target class, and the fault class of the target has no counterpart to align. Partial domain adaptation, in turn, addresses targets whose label space is a subset of the source label space by down-weighting source classes that are absent from the target (Cao et al., 2018). Our training setting resembles this case, because the target contributes only the normal class, but differs at deployment: the class that is absent during adaptation, the target fault, is exactly the class the detector must recognize later. Robustness-oriented work such as MAEFF, which fuses multiple sensors under missing observations (Yan et al., 2026), is complementary, since it addresses incomplete inputs within a labeled domain rather than the absence of target fault labels. Finally, active learning can reduce labeling effort under label scarcity in prognostics (Chen et al., 2025), but it presupposes that informative target samples exist to be queried, which is not the case before the first target fault.

Knowledge distillation (KD) offers a complementary route, in which a teacher model guides a student through soft probability outputs, transferring structural information about decision boundaries rather than raw features (Hinton et al., 2015). This is especially useful under class imbalance and in continual-learning settings, where the student inherits fault-detection capability while resisting catastrophic forgetting (F. Li et al., 2021; Y. Li et al., 2025; Z. Shi et al., 2022). Federated and collaborative learning have likewise been proposed to pool knowledge across machines without sharing raw data (Kairouz et al., 2021; L. Li et al., 2020; Sun et al., 2021; Wang et al., 2022), and are promising for cross-machine diagnosis where a well-instrumented source can support detection on a less-observed target. Their objective, however, is distributed or privacy-preserving optimization, not detection under a complete absence of target fault data.

Across these paradigms, several limitations persist for the problem we target. First, most methods implicitly assume that the data-scarce asset provides both normal and faulty samples, an assumption that frequently fails in practice, so that zero-shot detection from normal target data alone remains open. Second, aligning feature distributions does not by itself calibrate the classifier's decision boundary to the target. Standard domain-adversarial and discrepancy-based objectives align marginal feature distributions, and when the class proportions of the two domains differ, marginal alignment can even pull source features of one class onto target features of another (H. Zhao et al., 2019). Third, early-stage fault detection is still underexplored under realistic noise and operational variability.

These gaps motivate the framework in Sect. 3, which performs zero-shot fault detection from normal target data alone, anchors the decision boundary to the target via the proposed ACD loss, analyzes the resulting equilibrium under a normal-only target, sharpens incipient-fault sensitivity through an ahead-of-time labeling formulation, and uses SPC-based labeling to reduce manual annotation on the source.

## 3 Research Methodology

### 3.1 Test Rig and Data

This study uses the rolling bearing fatigue test rig and run-to-failure data described by Fan et al. (2024, 2025) (Fig. 1). The data contain run-to-failure experiments for two types of deep groove ball bearings, 6210 and 6011. The test rig comprises five main subsystems: a power system, a control system, a test specimen system, a drive system, and a loading system. A 3 kW electric motor, capable of reaching a maximum speed of 6000 r/min, drives the system. Rotational speed and other operational parameters are configured via a control computer, which regulates the motor speed through a frequency converter.

The test specimen system, shown in Fig. 1(b), consists of a test chamber, three bearing pedestals, four test bearings, a drive shaft, two shaft sleeves, and two tightening nuts. Four deep groove ball bearings are assembled onto the drive shaft in sequence, designated as Bearings 1 to 4 from left to right. These are installed into the three bearing pedestals: Pedestal 1 accommodates Bearing 1, Pedestal 2 accommodates Bearings 2 and 3, and Pedestal 3 accommodates Bearing 4. The assembled unit is then lowered into the test chamber. The loading system, depicted schematically in Fig. 1(c), applies radial and axial loads by adjusting the mass of the weights, thereby simulating different load conditions.

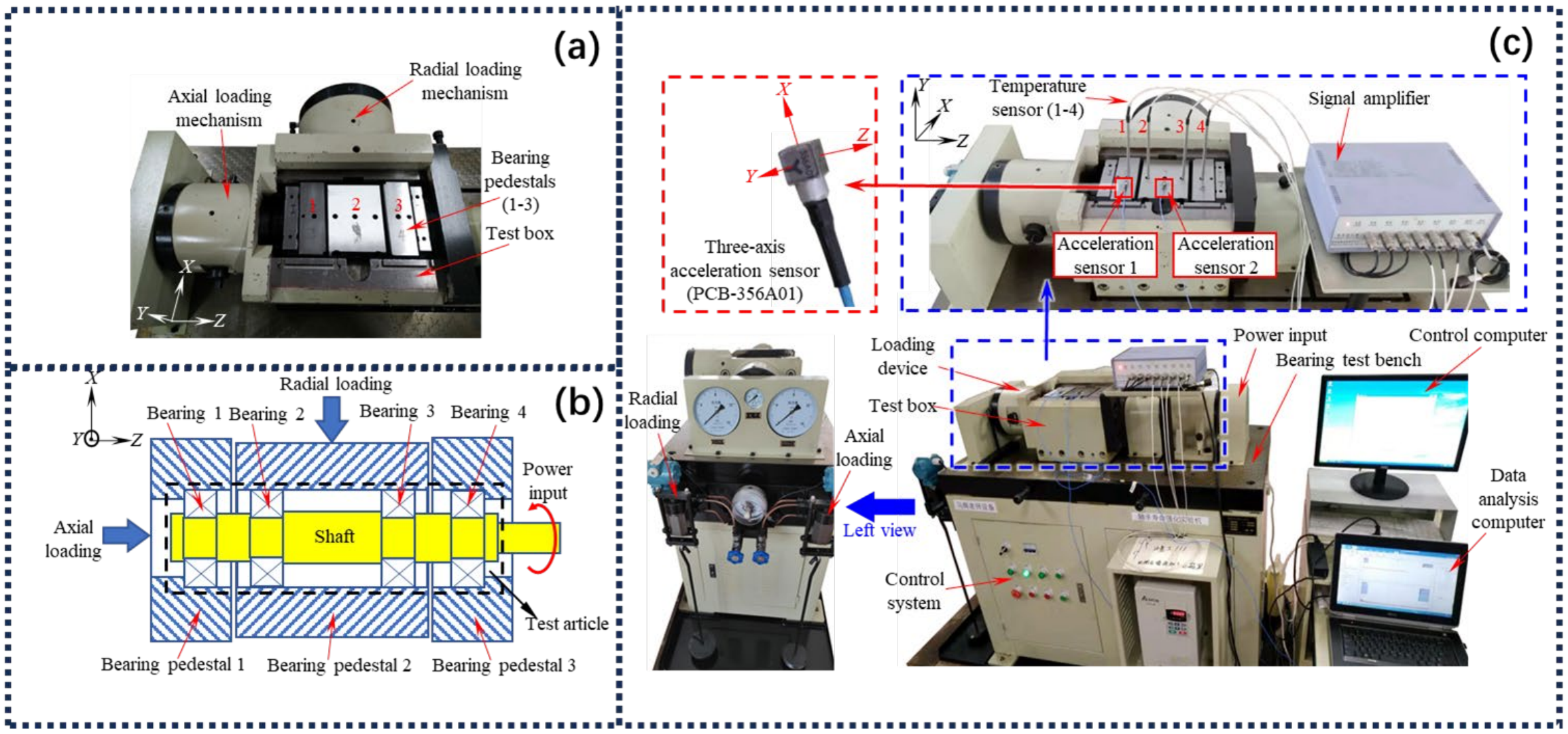

**Fig. 1** Rolling bearing fatigue test rig: (a) test rig, (b) photograph of the test specimen system, and (c) schematic diagram of the loading mechanism. Reproduced from Fan et al. (2024, 2025)

Both the 6210 and the 6011 experiments used four healthy deep groove ball bearings. The specifications are provided in Table 1. For the 6210 experiment, the rig ran at a constant rotational speed of 4200 r/min with a radial load of 6.8 kN and an axial load of 1.4 kN. Under these conditions, the bearing operated for approximately 1017 s until an automatic shutdown was triggered by a detected fault. For the 6011 experiment, the rig ran at the same rotational speed of 4200 r/min, with a radial load of 5.1 kN and an axial load of 1.4 kN. The test ran for approximately 3738 s.

***Table 1*** *Specifications of the bearing experiments*

| Parameter | | |
|---|---|---|
| Bearing Type | 6210 | 6011 |
| Inner Diameter | 50 mm | 55 mm |
| Outer Diameter | 90 mm | 90 mm |
| Width | 20 mm | 18 mm |
| Number of Balls | 10 | 10 |

Post-test inspection of the 6210 bearing, shown in Fig. 2 (Fan et al., 2024, 2025), revealed severe cage failure, with six of the ten connecting rivets on the cage fractured and three balls displaced from their original positions. This failure mode develops progressively, which makes early warning before the final failure both feasible and valuable for maintenance.

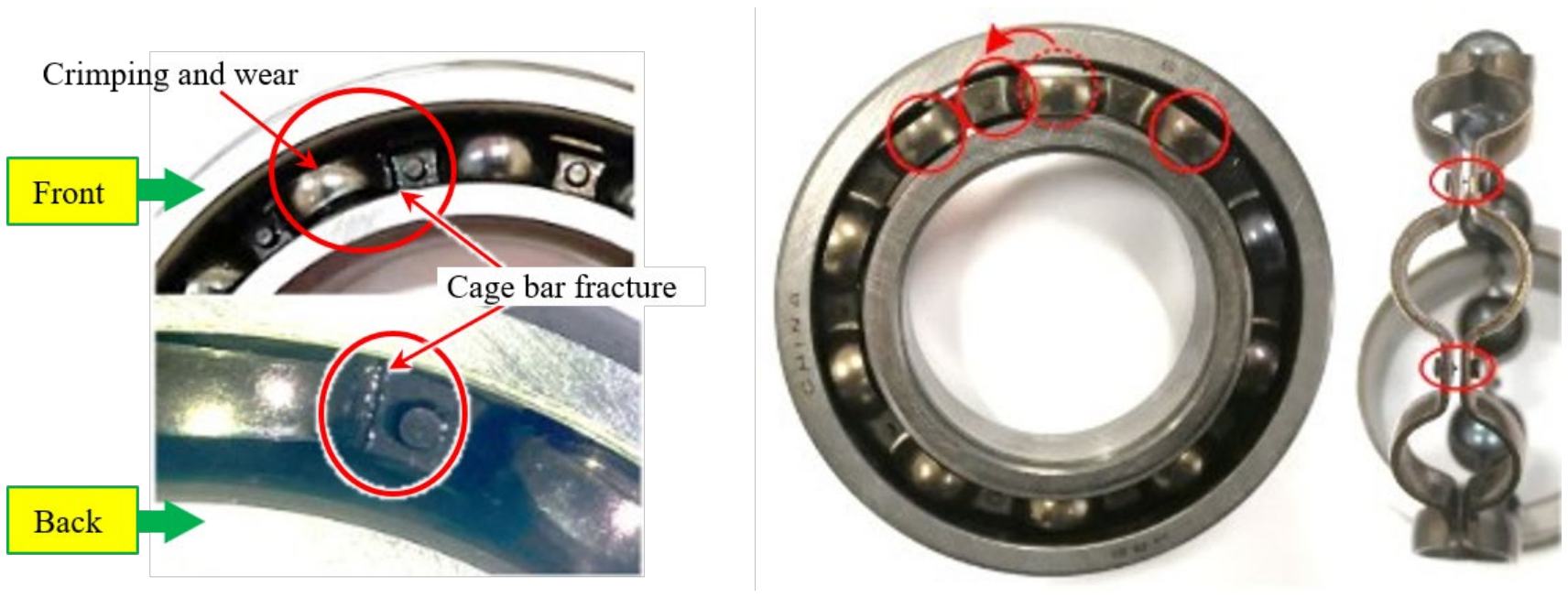

**Fig. 2** Post-test inspection of the 6210 bearing. Reproduced from Fan et al. (2024, 2025)

Vibration signals were acquired with two PCB-356A01 triaxial accelerometers (Fan et al., 2024, 2025). Accelerometer 1 was mounted on the exterior of Bearing Pedestal 1 and acquired vibration data from Bearing 1. Accelerometer 2 was mounted on the exterior of Bearing Pedestal 2 and acquired vibration data from Bearings 2 and 3. Accelerometer 1 measures acceleration in the Z direction and Accelerometer 2 in the Y and Z directions, where Y denotes the radial direction perpendicular to the shaft axis and Z denotes the radial direction parallel to the shaft axis, which yields three vibration channels. The signals were transmitted to a data acquisition unit, which converted the analog signals to digital format at a sampling frequency of 20 kHz, and were stored on a data analysis computer.

The resulting cross-asset early detection setting is as follows. Vibration signals acquired from the source bearing are first used to learn a diagnostic model that encodes the discriminative structure separating normal from degraded operation. The knowledge transfer architecture of Sect. 3.4 then adapts this model to the target bearing, using only the target's early-stage normal data to align the two assets' feature representations and to anchor the decision boundary. The adapted model is deployed on the target bearing's incoming signal stream to issue early warnings that can inform maintenance decisions for the target.

To demonstrate the cross-asset knowledge transfer architecture, the 6210 experiment is designated as the source domain (Bearing A), providing both normal and faulty data with automatically generated labels, and the 6011 experiment is designated as the target domain (Bearing B). Because the two bearings differ in type and in radial load, the transfer involves both an asset shift and an operating-condition shift, although both experiments were performed on the same rig at the same speed.

### 3.2 Problem Formulation

In practical industrial settings, a target bearing may operate under normal conditions for an extended period, while fault-state data are rarely observed or recorded. Conversely, historical vibration data containing both normal and degraded periods are often available from an analogous source bearing (Bearing A). This setting motivates transferring diagnostic knowledge from the data-rich source bearing to the data-scarce target bearing (Bearing B).

Let $x_t^A \in \mathbb{R}^C$ denote the $C$-channel vibration measurement at time index $t$ from Bearing A, for $t = 1, \dots, T_A$. Similarly, $x_t^B \in \mathbb{R}^C$ denotes the measurement from Bearing B, for $t = 1, \dots, T_B$. We assume that a stage indicator sequence $\{s_t^A\}_{t=1}^{T_A}$ (and $\{s_t^B\}_{t=1}^{T_B}$ for evaluation) is available or can be derived, for example from control-chart statistics (Sect. 3.3). A time index $t$ is considered abnormal if $s_t \geq \zeta$, where $\zeta$ is a predefined abnormality threshold.

To enable early warning, we construct sliding-window samples. Given window length $W$, stride $S$, and prediction horizon $H$ (all in sampling points), each sample is indexed by a window start time $i \in \{1, 1+S, 1+2S, \dots, T-W-H+1\}$, and the corresponding input is the windowed multichannel segment

$$X_i = [x_i, x_{i+1}, \dots, x_{i+W-1}] \in \mathbb{R}^{C \times W}. \qquad (1)$$

The early-warning label of this window is defined by whether the abnormal stage occurs more than $\eta$ times within a future interval that starts $H$ sampling points after the window start:

$$y_i = \mathbb{I}\left( \sum_{t=i+H}^{i+H+W-1} \mathbb{I}\,(s_t \geq \zeta) > \eta \right), \qquad (2)$$

that is, $y_i = 1$ if more than $\eta$ of the time indices in $[i+H,\, i+H+W-1]$ are abnormal, and $y_i = 0$ otherwise. This definition matches the operational goal of using evidence observed in the current window to predict whether an abnormal state will be present $H$ sampling points later. In maintenance terms, $H$ defines the future labeling interval relative to the window start; the realized lead time depends on the timing of the model's alarms.

In the target domain, only a limited amount of early-stage normal data is available for adaptation. Specifically, we assume that for Bearing B we only observe windows whose starts lie in an early normal period, $\mathcal{D}_B^T = \{(X_i^B, y_i^B = 0)\}$ with window starts in $\{1, \dots, M\}$, where $M$ is the number of time points observed before any fault appears. We denote the full data of Bearing B by $\mathcal{D}_B$ and the data on which predictions are required by $\mathcal{D}_B^C$, the windows starting after time $M$. We write $N_A$ and $N_T$ for the numbers of source windows and target adaptation windows, respectively. Our objective is to learn a classifier $f(\cdot)$ that predicts the probability of an upcoming abnormal state,

$$p_i = f(X_i^B) \in [0,1], \qquad \hat{y}_i = \mathbb{I}(p_i \geq \tau), \tag{3}$$

and to accurately predict the early-warning labels of the later windows in $\mathcal{D}_B^C$, beyond the early normal period.

The proposed framework proceeds in three phases. In Phase 1, unsupervised label identification based on a control chart determines the labels of the source domain data and yields the labeled source set $\mathcal{D}_A$ (Sect. 3.3). In Phase 2, the Adapted Domain-Adversarial Neural Network (ADANN) transfers knowledge from $\mathcal{D}_A$ to Bearing B by incorporating $\mathcal{D}_B^T$ in the training procedure (Sect. 3.4). In Phase 3, the trained model performs online early fault detection on the target (Sect. 3.5).

### 3.3 Unsupervised Label Identification on the Source Asset

Historical data from Bearing A can be abundant, raw, and devoid of annotations. Manually labeling extensive time series spanning months or years is labor-intensive, subjective, and impractical for real-world deployment. Furthermore, manual labeling often fails to capture the precise point of fault initiation, labeling broad segments as faulty and potentially obscuring the subtle, incipient patterns that are critical for early detection. To reduce this burden and construct a consistent source dataset, an SPC method, specifically the exponentially weighted moving average (EWMA) control chart, is employed to identify the states of Bearing A. In this study, the root mean square (RMS) of the vibration signal is monitored. For the sequence of windowed segments from Bearing A, the RMS values are computed channel by channel, yielding a time series $r_t$, $t = 1,2, \dots, T$, where $r_t$ is the RMS value at time interval $t$. The EWMA statistic $z_t$ is defined recursively as

$$z_t = \lambda r_t + (1-\lambda) z_{t-1}, \qquad t = 1,2,\dots,T, \qquad z_0 = \mu_0, \tag{4}$$

where $\lambda \in (0,1]$ is a smoothing constant that determines the memory of the chart.

The upper control limit is constructed from the mean $\mu_0$ and standard deviation $\sigma_0$ of the RMS series over an initial normal operating period,

$$\mathrm{UCL}_t = \mu_0 + L\sigma_0 \sqrt{\frac{\lambda}{2-\lambda}\left[1-(1-\lambda)^{2t}\right]}, \tag{5}$$

where $L$ is the control limit width. The labeling rule is as follows. As long as $z_t$ remains below the upper control limit, the bearing at time $t$ is considered normal ($s_t^A = 0$). The first time $z_t$ exceeds $\mathrm{UCL}_t$, the process is deemed out of control, indicating the initiation of a fault, and all subsequent time points are labeled abnormal ($s_t^A = 1$). The window labels of the source then follow from Eq. (2) with $\zeta = 1$; because Eq. (2) looks $H$ sampling points ahead, the source labels are shifted forward by the prediction horizon, so that each source window is labeled with the state reached $H$ sampling points later. This procedure yields the labeled source dataset $\mathcal{D}_A = \{(X_i^A, y_i^A)\}_{i=1}^{N_A}$.

The control chart may not detect the fault onset perfectly and may raise false alarms, which could mislead the subsequent transfer step. Therefore, the segments flagged as anomalous are manually verified in this study. Even so, verifying the flagged segments requires substantially less effort than labeling the entire record manually.

The verified source dataset $\mathcal{D}_A$ serves as the input for the next stage of the framework. The following subsection details how the framework leverages $\mathcal{D}_A$, together with the normal target data $\mathcal{D}_B^T$, to learn a feature representation that is discriminative for fault detection and transferable across bearings.

### 3.4 Adapted Domain-Adversarial Neural Network (ADANN)

Following the acquisition of a labeled source dataset, the next step is to transfer the diagnostic knowledge embedded in these data to the target bearing. Knowledge transfer is required because the distributions of vibration data collected from different bearings, even of the same type, often exhibit a domain shift due to variations in manufacturing tolerances, installation conditions, and operating environments. Domain-Adversarial Neural Networks (DANN) (Ganin et al., 2016) mitigate this shift by learning feature representations that are discriminative on the source domain and invariant to the differences between the source and target domains. The invariance is achieved through adversarial training, which encourages the feature extractor to confuse a domain classifier and thereby aligns the feature distributions of the two domains.

As shown in Fig. 3, the DANN architecture comprises three trainable components: a feature extractor $G_f$, a label predictor $G_y$, and a domain classifier $G_d$. The feature extractor, parameterized by $\theta_f$, maps input windows from both domains into a shared feature space. The label predictor, parameterized by $\theta_y$, takes the features of source windows and predicts their labels by minimizing the binary cross-entropy

$$\mathcal{L}_y(\mathcal{D}_A;\theta_f,\theta_y) = -\frac{1}{N_A}\sum_{i=1}^{N_A}\left[y_i^A \ln G_y\left(G_f(X_i^A)\right) + (1-y_i^A)\ln\left(1-G_y\left(G_f(X_i^A)\right)\right)\right]. \quad (6)$$

The domain classifier, parameterized by $\theta_d$, outputs the probability that a feature originates from the source domain and is trained by minimizing the domain classification loss

$$\mathcal{L}_d(\mathcal{D}_A,\mathcal{D}_B^T;\theta_f,\theta_d) = -\frac{1}{N_A}\sum_{i=1}^{N_A}\ln G_d\left(G_f(X_i^A)\right) - \frac{1}{N_T}\sum_{X\in\mathcal{D}_B^T}\ln\left(1-G_d\left(G_f(X)\right)\right). \quad (7)$$

A gradient reversal layer (GRL) is positioned between the feature extractor and the domain classifier. In the forward pass the GRL acts as the identity; in backpropagation it reverses the sign of the gradients flowing from the domain classifier to the feature extractor. As a result, the domain classifier is trained to minimize $\mathcal{L}_d$ while the feature extractor is trained to maximize it, which yields features that the domain classifier cannot distinguish. The standard DANN objective is

$$\min_{\theta_f,\theta_y}\max_{\theta_d} \mathcal{E}(\theta_f,\theta_y,\theta_d) = \mathcal{L}_y(\mathcal{D}_A;\theta_f,\theta_y) - \alpha\mathcal{L}_d(\mathcal{D}_A,\mathcal{D}_B^T;\theta_f,\theta_d), \quad (8)$$

where $\alpha$ weights the domain classification loss. At convergence, the label information of Bearing A is encoded in $G_y$, the domain-invariant representation is encoded in $G_f$ through the adversarial signal of $G_d$, and the composition $G_y \circ G_f$ is applied to predict the labels of $\mathcal{D}_B^C$.

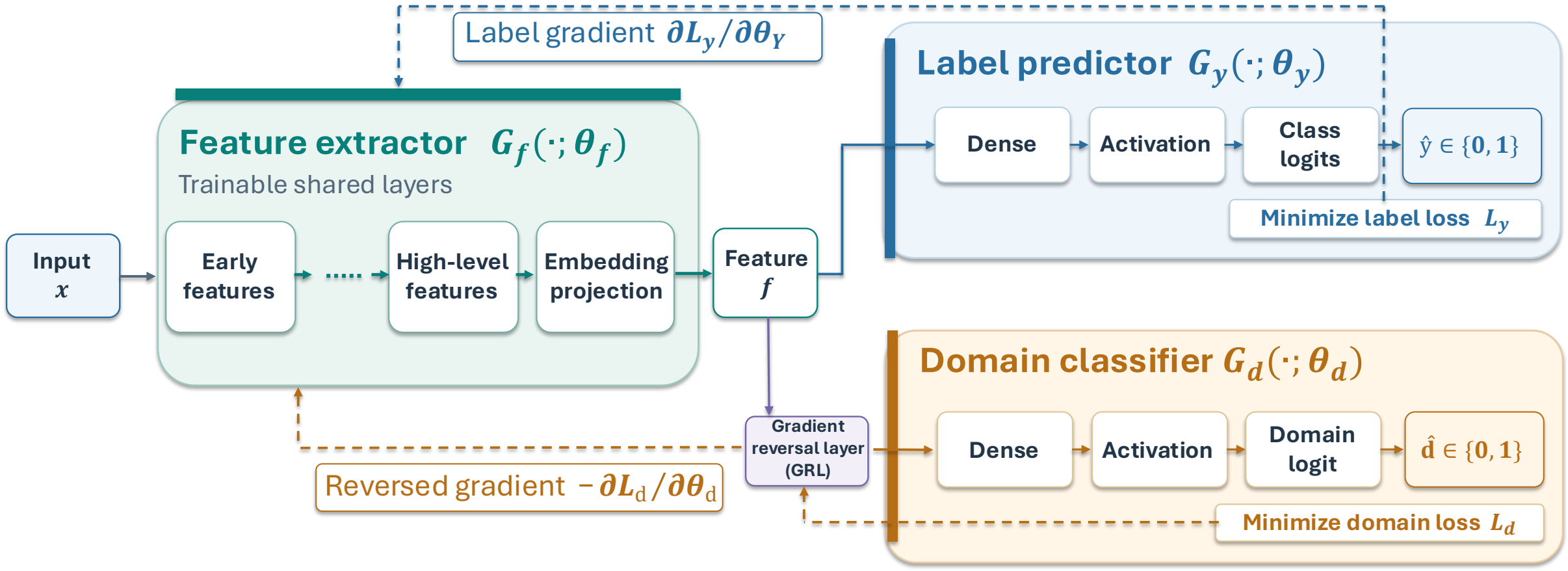

**Fig. 3** Architecture of the domain-adversarial neural network. Reproduced from Ganin et al. (2016)

The standard DANN objective only makes the domains indistinguishable. It does not use the limited but valuable information available in the target's normal data, and it does not enforce a finer alignment of the distributions. To improve transfer for early fault detection, ADANN incorporates two modifications. First, a Maximum Mean Discrepancy (MMD) loss (Gretton et al., 2006) explicitly minimizes the distributional distance between the source and target features. Second, an Adapted Classification-Domain (ACD) loss involves the target's normal data in the classification objective and thereby anchors the decision boundary on the target bearing.

### *3.4.1 Maximum Mean Discrepancy Loss*

Adversarial training is effective at achieving global domain alignment, but it may not capture finer, higher-order statistical differences between the two domains. To complement the adversarial mechanism with a direct and stable measure of distributional similarity, an MMD loss is incorporated into ADANN. MMD is a nonparametric distance between two probability distributions based on their embeddings in a reproducing kernel Hilbert space (RKHS) $\mathcal{H}$; with a characteristic kernel, the mean embeddings of two distributions coincide if and only if the distributions are identical.

Let $p_s$ and $p_t$ denote the distributions of the source and target features produced by $G_f$. Given source features $F_A = \{f_i^A\}_{i=1}^{N_A}$ drawn from $p_s$ and target features $F_B = \{f_j^B\}_{j=1}^{N_T}$ drawn from $p_t$, the squared MMD between the two empirical distributions is

$$\text{MMD}^2(F_A, F_B) = \left\| \frac{1}{N_A} \sum_{i=1}^{N_A} \phi\left(f_i^A\right) - \frac{1}{N_T} \sum_{j=1}^{N_T} \phi\left(f_j^B\right) \right\|_{\mathcal{H}}^2, \tag{9}$$

where $\phi(\cdot)$ is the feature map associated with a kernel $k(\cdot,\cdot)$, such that $k(f, f') = \langle \phi(f), \phi(f') \rangle_{\mathcal{H}}$. We employ the radial basis function (RBF) kernel $k(f, f') = \exp\left(-\| f - f' \|^2 / (2\sigma^2)\right)$ with bandwidth $\sigma$. The MMD loss $\mathcal{L}_{mmd}$ is computed as this squared distance on each mini-batch of source and target features. Adding it to the training objective penalizes the feature extractor whenever the source and target feature distributions diverge, which provides a complementary supervisory signal to the adversarial process. The objective becomes

$$\min_{\theta_f, \theta_y} \max_{\theta_d} \mathcal{E} = \mathcal{L}_y\left(\mathcal{D}_A; \theta_f, \theta_y\right) - \alpha \mathcal{L}_d\left(\mathcal{D}_A, \mathcal{D}_B^T; \theta_f, \theta_d\right) + \mathcal{L}_{mmd}\left(\mathcal{D}_A, \mathcal{D}_B^T; \theta_f\right). \tag{10}$$

The adversarial term and the MMD term both act on the marginal feature distributions. Neither of them evaluates the label predictor $G_y$ on target data: $G_y$ remains trained exclusively on source-labeled windows, which leaves the decision boundary uncalibrated for the target's normal condition. The ACD loss addresses this limitation.

### *3.4.2 Adapted Classification-Domain Loss*

Trained on labeled source data only, the label predictor learns a decision boundary that separates normal and faulty conditions within the source feature distribution. After alignment, the target features lie close to the source distribution, but the part of the feature space occupied by the target's normal windows need not coincide with the region the source-trained boundary labels as normal. Target normal windows that fall on the wrong side of this boundary produce false alarms. To address this, the ACD loss extends the classification objective across the two domains.

The classification objective of ADANN consists of two parts. The first is the source classification loss $\mathcal{L}_y$ of Eq. (6), which preserves the ability of $G_y$ to discriminate normal from faulty states on the labeled source data. The second is a target normality term, which feeds the known normal target windows $\mathcal{D}_B^T$ through $G_f$ and $G_y$ and penalizes any probability of the abnormal class:

$$\mathcal{L}_{ACD}\left(\mathcal{D}_B^T;\theta_f,\theta_y\right) = -\frac{1}{N_T}\sum_{X\in\mathcal{D}_B^T} \ln\left(1 - G_y\left(G_f(X)\right)\right). \quad (11)$$

We refer to Eq. (11) as the ACD loss, since it adapts the classification objective to the target domain. The complete ADANN objective is

$$\min_{\theta_f,\theta_y}\max_{\theta_d} \mathcal{E} = \mathcal{L}_y\left(\mathcal{D}_A;\theta_f,\theta_y\right) + \mathcal{L}_{ACD}\left(\mathcal{D}_B^T;\theta_f,\theta_y\right)$$

$$-\alpha\mathcal{L}_d\left(\mathcal{D}_A,\mathcal{D}_B^T;\theta_f,\theta_d\right) + \mathcal{L}_{mmd}\left(\mathcal{D}_A,\mathcal{D}_B^T;\theta_f\right). \quad (12)$$

With the ACD loss, the label predictor is no longer a passive recipient of aligned features. It is guided to place the target's normal windows inside its normal region, so that the definition of normal is anchored to Bearing B's own normal operating condition. Whether incipient fault signals of Bearing B enter the source-trained fault region is evaluated empirically, because the ACD term contains no target fault examples. The ACD loss is the only term in Eq. (12) that evaluates the label predictor on target data; Sect. 3.4.3 makes its role precise. The training procedure of ADANN is summarized in Algorithm 1.

**Algorithm 1** Training Procedure for the Adapted Domain-Adversarial Neural Network (ADANN)

**Input**: labeled source dataset $\mathcal{D}_A$ (Bearing A); normal-only target dataset $\mathcal{D}_B^T$ (Bearing B); number of epochs $E$; batch size $B$; domain weight $\alpha$; kernel bandwidth $\sigma$

**Initialize** parameters $\theta_f$ of $G_f$, $\theta_y$ of $G_y$, and $\theta_d$ of $G_d$

**For** epoch $e$ **from** 1 **to** $E$:

**For** each iteration of the epoch:

**Step 1:** Sample a source mini-batch $(X_s, Y_s)$ of size $B$ from $\mathcal{D}_A$

**Step 2:** Sample a target mini-batch $X_t$ of size $B$ from $\mathcal{D}_B^T$ (all labels normal)

**Step 3:** Compute features $F_s = G_f(X_s)$ and $F_t = G_f(X_t)$

**Step 4:** Compute $\mathcal{L}_y$ on $\left(G_y(F_s), Y_s\right)$ by Eq. (6), $\mathcal{L}_{ACD}$ on $G_y(F_t)$ by Eq. (11), $\mathcal{L}_d$ on $\left(G_d(F_s), G_d(F_t)\right)$ by Eq. (7), and $\mathcal{L}_{mmd} = \mathrm{MMD}^2(F_s, F_t)$ by Eq. (9)

**Step 5:** Compute $\mathcal{E}$ by Eq. (12)

**Step 6:** Update $\theta_f, \theta_y$ by descending $\mathcal{E}$ and $\theta_d$ by descending $\mathcal{L}_d$ (implemented jointly through the GRL) using the Adam optimizer

**Output:** Trained feature extractor $G_f$, label predictor $G_y$, and domain classifier $G_d$

### *3.4.3 Equilibrium of ADANN under a Normal-Only Target*

By the EWMA-based labeling of Sect. 3.3, the source domain $\mathcal{D}_A$ contains both normal and abnormal windows, whereas the target asset contributes only normal windows $\mathcal{D}_B^T$ to training. The source feature density induced by $G_f$ is therefore a class mixture,

$$p_s(f) = \pi_N\, p_s^N(f) + \pi_A\, p_s^A(f), \qquad \pi_N + \pi_A = 1, \tag{13}$$

where $p_s^N$ and $p_s^A$ are the class-conditional feature densities of the source normal and abnormal windows, $\pi_A$ is the abnormal proportion fixed by the control-chart labels, and the target density $p_t$ is purely normal. Both alignment mechanisms operate on these marginals: the domain classifier separates source features drawn from the mixture $p_s$ from target features drawn from $p_t$ , and $\mathcal{L}_{mmd}$ is computed between the same two marginals. The following theorem characterizes the domain-adversarial component of Eq. (12) under the classification correctness that ADANN must preserve at deployment.

**Theorem 1.** *Assume that (i) $G_f$, $G_y$, and $G_d$ have sufficient capacity; (ii) at each training step, $G_d$ attains the minimum of $\mathcal{L}_d$ for the current $G_f$; and (iii) the label predictor $G_y$ classifies the source windows and the target normal windows correctly, the property that $\mathcal{L}_y$ and $\mathcal{L}_{ACD}$ drive toward at convergence. Then the domain-adversarial component of Eq. (12) at the optimal domain classifier $D^*$ satisfies*

$$-\alpha\mathcal{L}_d(D^*) \geq -\alpha\ln 4 + 2\alpha\,\delta(\pi_A), \tag{14}$$

*where*

$$\delta(\pi_A) = \ln 2 - \frac{2-\pi_A}{2}\, H\left(\frac{1-\pi_A}{2-\pi_A}\right) = H\left(\frac{\pi_A}{2}\right) - \frac{1}{2}H(\pi_A), \tag{15}$$

*with $H(\cdot)$ the binary entropy in nats. Equality in Eq. (14) holds if and only if the source normal features and the target normal features coincide, $p_s^N(f) = p_t(f)$ for almost every $f$, while the source abnormal features remain separated from them. At this aligned state, the optimal domain classifier outputs 1 on the abnormal features and $\pi_N/(1+\pi_N)$ on the normal ones, and the MMD term equals its residual value $\mathcal{L}_{mmd}^{\dagger} = \pi_A^2 \parallel \mu_s^A - \mu_s^N \parallel_{\mathcal{H}}^2$, where $\mu_s^N$ and $\mu_s^A$ are the class-conditional kernel mean embeddings. Moreover, $\delta$ is continuous and strictly increasing on $[0,1]$ with $\delta(0) = 0$ and $\delta(1) = \ln 2$; hence, as $\pi_A \to 0$, Eq. (14) reduces to the constant $-\alpha\ln 4$ attained at $p_s = p_t$, the classical equilibrium of domain-adversarial training.*

The proof is given in Appendix A. Theorem 1 shows that, under a normal-only target, the domain-adversarial component cannot reach the classical value $-\alpha\ln 4$: the abnormal proportion $\pi_A$ of the source has no counterpart in $\mathcal{D}_B^T$, and the classification losses keep it on the abnormal side of the boundary, so the gap $2\alpha\,\delta(\pi_A)$ quantifies the irreducible mismatch of adapting to a normal-only target. The bound is attained exactly at the state expected of successful adaptation, in which the normal clusters of the two bearings merge ($p_s^N = p_t$) and the abnormal cluster remains separated. Without the classification terms, the adversarial and MMD terms would instead be minimized at $p_s = p_t$, which drags source abnormal features onto the target normal data. The classification terms in Eq. (12) penalize this failure mode of marginal alignment under label shift, and Theorem 1 characterizes the outcome only when its exact-classification assumption holds.

The theorem also yields a diagnostic for training. At the aligned state, the domain classifier outputs approach $\pi_N/(1+\pi_N)$ on target windows and 1 on source abnormal windows, the domain classification loss approaches $\ln 4 - 2\delta(\pi_A)$, and $\mathcal{L}_{mmd}$ plateaus at $\mathcal{L}_{mmd}^{\dagger}$ rather than at zero. Under the theorem's exact-classification and aligned-state assumptions, a nonzero residual is therefore compatible with the normal-only target setting. Theorem 1 is a population-level statement under assumptions (i) to (iii); with finite mini-batches and an adversary that is only approximately optimal, these values are approached rather than attained.

**Remark 1 (role of the ACD loss).** Assumption (iii) contains three conditions (Eq. (A.2) in Appendix A). Two of them, $p_s^A(R_A) = 1$ and $p_s^N(R_A) = 0$ for the abnormal region $R_A$ of the classifier, concern the source domain and are what $\mathcal{L}_y$ drives toward. The third, $p_t(R_A) = 0$, requires that the target's normal features lie outside the abnormal region; it concerns only the target, and among the terms of Eq. (12) only $\mathcal{L}_{ACD}$ evaluates $G_y$ on target data. The ACD loss controls this condition directly. Since $\mathbb{I}(p \geq \tau) \leq \ln\frac{1}{1-p}/\ln\frac{1}{1-\tau}$ for every $p \in [0,1)$, the empirical false-alarm rate on the adaptation windows satisfies

$$\widehat{\mathrm{FAR}}_T := \frac{1}{N_T}\sum_{X\in\mathcal{D}_B^T}\mathbb{I}\left(G_y\left(G_f(X)\right) \geq \tau\right) \leq \frac{\mathcal{L}_{ACD}}{\ln\bigl(1/(1-\tau)\bigr)}, \tag{16}$$

so that, for $\tau = 0.5$, $\widehat{\mathrm{FAR}}_T \leq \mathcal{L}_{ACD}/\ln 2$. Equation (16) is an in-sample statement about the adaptation windows. It is not a generalization bound: extending it to future normal windows of the target would require a held-out normal interval or additional stationarity, capacity, and concentration assumptions, none of which is claimed here. No analogous target-side control exists for missed detections. With zero target fault samples, detection of target faults relies on the source abnormal region $R_A$ and on the alignment $p_s^N \approx p_t$, and the distribution of target fault features cannot be estimated from $\mathcal{D}_B^T$. ADANN is therefore expected to reduce false alarms relative to source-only DANN and MMD alignment, without a comparable guarantee on recall; Sect. 4.2 examines this prediction empirically.

**Remark 2 (relation to existing results).** Lemmas 1 and 2 in Appendix A follow the classical analysis of adversarial training (Goodfellow et al., 2014). The quantity $\delta(\pi_A)$ equals the Jensen-Shannon divergence between the label distributions of the two domains, $\mathrm{Bernoulli}(\pi_A)$ and $\mathrm{Bernoulli}(0)$, so the lower bound in Eq. (14) is an instance of the label-shift argument of H. Zhao et al. (2019) specialized to a normal-only target. What Theorem 1 adds for the present setting is the exact attainment condition $p_s^N = p_t$, the equilibrium outputs of the domain classifier, and the MMD residual, which together identify the aligned state that ADANN is designed to reach.

### 3.5 Overall Framework for Zero-Shot Cross-Asset Fault Detection

Building on the components of the preceding subsections, Fig. 4 illustrates the architecture and workflow of the proposed zero-shot cross-asset fault detection framework. The framework operates in three phases: offline source labeling, offline ADANN training, and online early fault detection on the target.

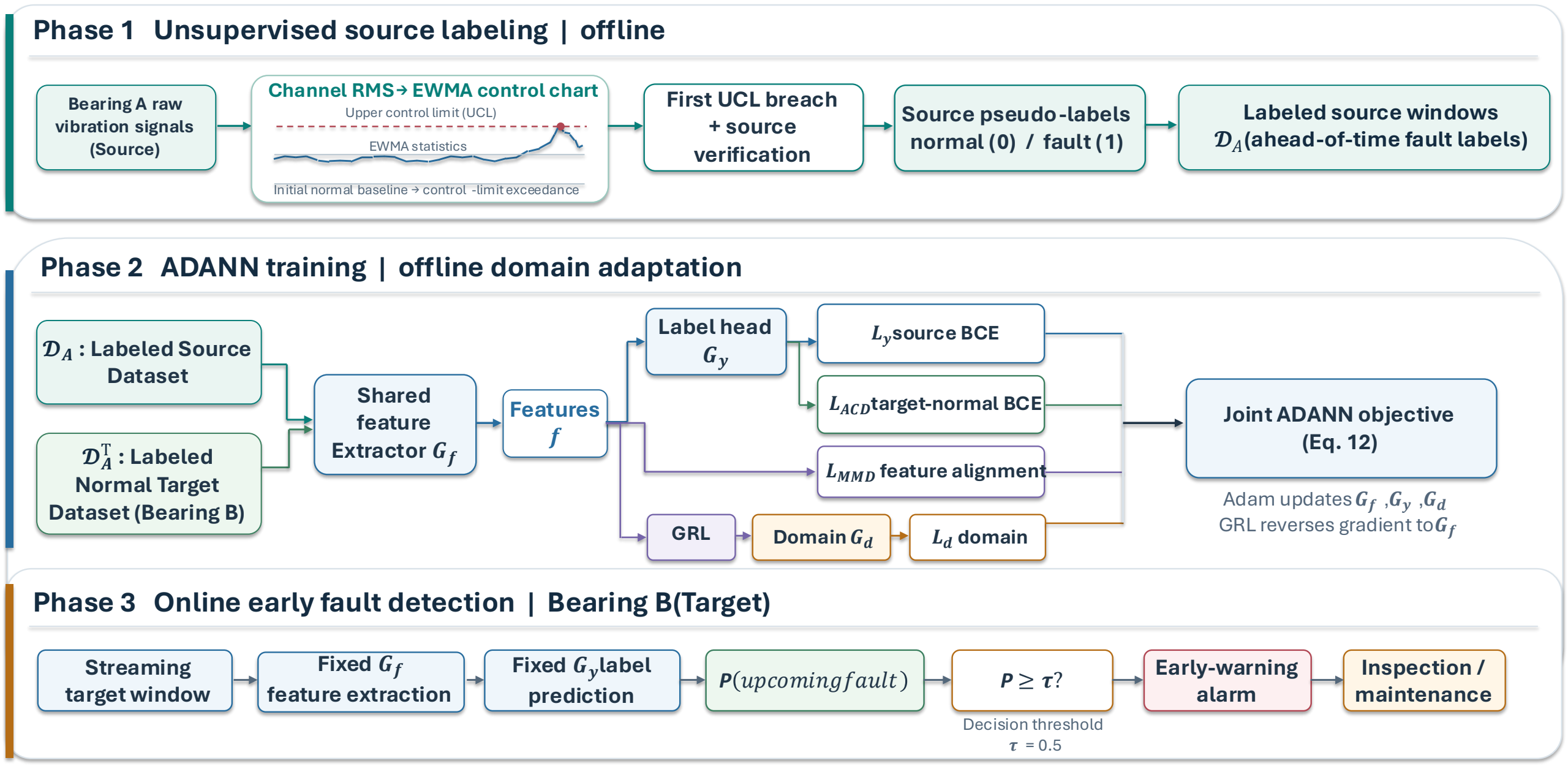

**Fig. 4** Overview of the proposed framework. Unlabeled vibration data from the source bearing (Bearing A) are labeled by an EWMA control chart; the labeled source data and the early-stage normal data of the target bearing (Bearing B) are used to train ADANN; the trained model performs online early fault detection on Bearing B

The first phase, unsupervised label identification, processes the raw vibration signals from the source bearing. Condition indicators such as the RMS are extracted and monitored with an EWMA control chart whose limits are derived from an initial normal operating period, and each segment is labeled as normal or faulty. After verification of the flagged segments, this phase yields the labeled source dataset $\mathcal{D}_A$.

The second phase, ADANN training, uses $\mathcal{D}_A$ together with the normal target dataset $\mathcal{D}_B^T$. Source and target features are passed to the label predictor, which is trained with $\mathcal{L}_y$ on source windows and $\mathcal{L}_{ACD}$ on target windows, and, through the gradient reversal layer, to the domain classifier. The MMD loss is computed directly between the source and target features. The four loss components are combined into the objective of Eq. (12), which updates the parameters of all three networks by backpropagation.

The third phase deploys the trained model for online fault detection on Bearing B. Streaming vibration windows are passed through the fixed feature extractor $G_f$ and label predictor $G_y$, which outputs the probability of an upcoming abnormal state. Because the ACD loss anchors the target's normal condition in the normal region, a low predicted probability indicates normal operation, and a rise above the threshold $\tau$ issues an early warning. Since the labels are defined $H$ sampling points ahead of the window start, a warning can precede the labeled abnormal state. The realized maintenance lead time depends on when alarms occur relative to that state. The complete procedure is summarized in Algorithm 2 and validated in Sect. 4.

**Algorithm 2** Zero-shot cross-asset early fault detection framework

**Input**: raw source data $\{x_t^A\}_{t=1}^{T_A}$ (Bearing A); raw early-stage target data $\{x_t^B\}_{t=1}^{M}$ (Bearing B); streaming target data $\{x_t^B\}_{t>M}$; ADANN hyperparameters $E$, $B$, $\alpha$, $\sigma$; decision threshold $\tau$
**Step 1:** Window the early-stage target data by Eq. (1) and label all windows as normal to form $\mathcal{D}_B^T$
**Step 2:** Label the source data with the EWMA chart of Eqs. (4) and (5), verify the flagged segments, and form $\mathcal{D}_A$ by Eqs. (1) and (2)
**Step 3:** Train ADANN on $\mathcal{D}_A$ and $\mathcal{D}_B^T$ by Algorithm 1
**Step 4:** For each incoming target window $X_i^B$, compute $p_i = G_y\left(G_f(X_i^B)\right)$ and issue an alarm if $p_i \geq \tau$ (Eq. (3))
**Output:** early-warning predictions $\hat{y}_i$ for the target windows in $\mathcal{D}_B^C$

# 4 Case Study

This section evaluates the proposed methodology on the bearing data of Sect. 3.1. The setup is described in Sect. 4.1, the main results in Sect. 4.2, the implications for maintenance decisions in Sect. 4.3, and the robustness experiments in Sect. 4.4.

## 4.1 Experimental Setup

For the source domain (Bearing A), the last 40 s of the run-to-failure record were used, and the anomalous period begins at 32 s of this segment. At a sampling frequency of 20 kHz, the segment contains 800,000 time points on each of the three acceleration channels. The vibration signals of the analyzed segments of Bearing A and Bearing B are shown in Fig. 5.

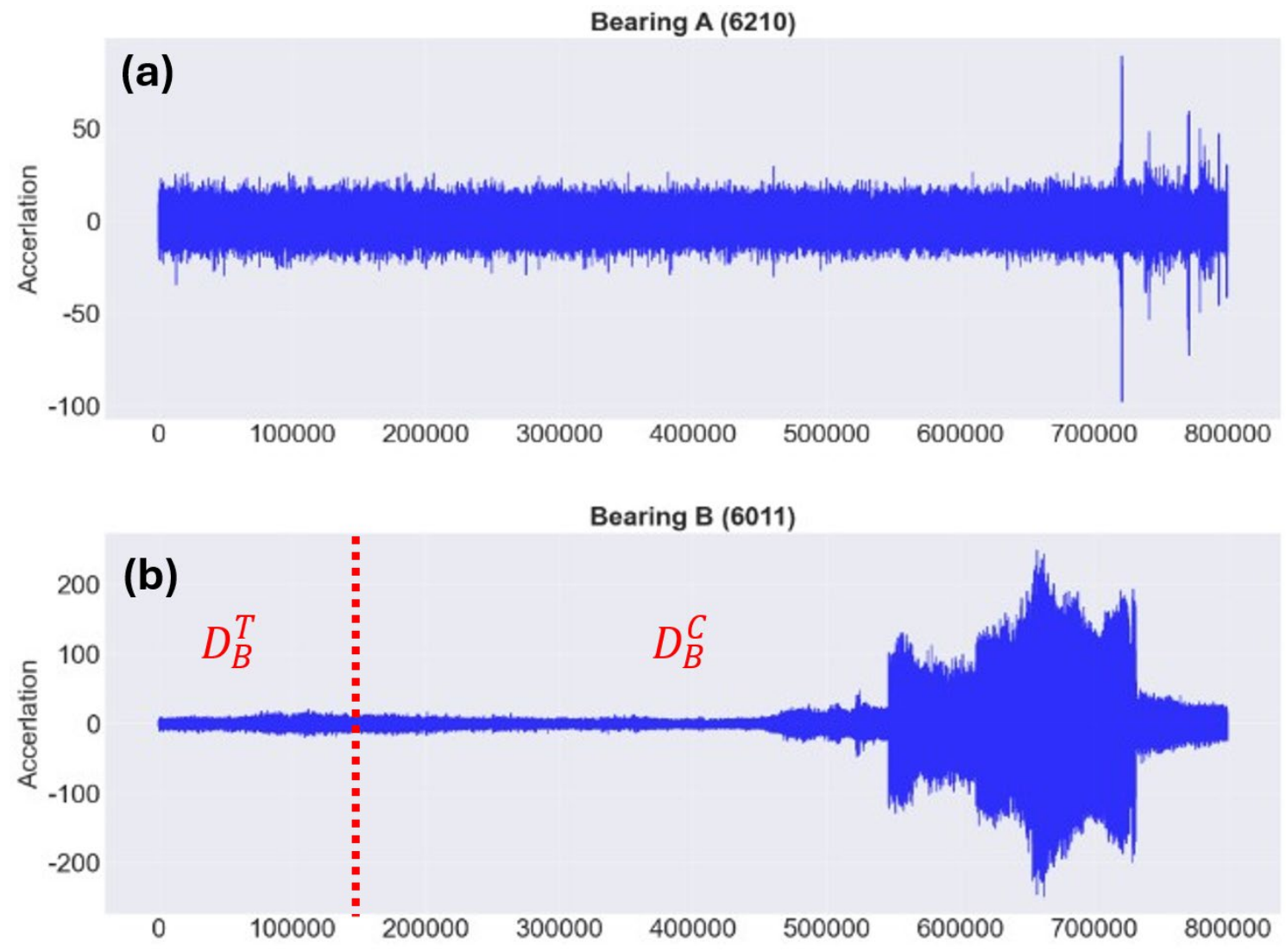

**Fig. 5** Vibration signal (Y direction) of the 40 s segments analyzed for (a) Bearing A (6210 experiment) and (b) Bearing B (6011 experiment). The dashed line in (b) separates the adaptation set $\mathcal{D}_B^T$ from the test set $\mathcal{D}_B^C$

For the target domain (Bearing B), the last 40 s of the record were also used, with the evaluation fault boundary set at 20 s of this segment; the evaluation stage indicator $s_t^B$ is set to 1 from this boundary onward. The target data are divided into two subsets: the adaptation set $\mathcal{D}_B^T$, consisting of the first 8 s of the segment in Fig. 5(b), and the test set $\mathcal{D}_B^C$, consisting of the remaining 32 s. Each window contains $W = 1000$ consecutive sampling points (0.05 s), the stride is $S = 200$, the prediction horizon is 2 s ($H = 40{,}000$ sampling points), and $\eta = 200$, so that a window is labeled abnormal if more than 20% of the corresponding future interval is abnormal. For the binary EWMA-derived indicator, $\zeta = 1$. For the unsupervised label identification of Bearing A, the EWMA chart was applied to the RMS series of the raw vibration signal with smoothing constant $\lambda = 0.3$ and control limit width $L = 3.6$.

ADANN comprises three networks: a feature extractor, a label predictor, and a domain classifier, whose architectures are shown in Fig. 6. The model was trained with the Adam optimizer at an initial learning rate of 0.001, a batch size of 128, and 100 epochs. The domain weight was set empirically to $\alpha = 1.0$, the ACD and MMD terms enter Eq. (12) with unit weight, and the RBF kernel bandwidth was $\sigma = 1.0$. The decision threshold is $\tau = 0.5$.

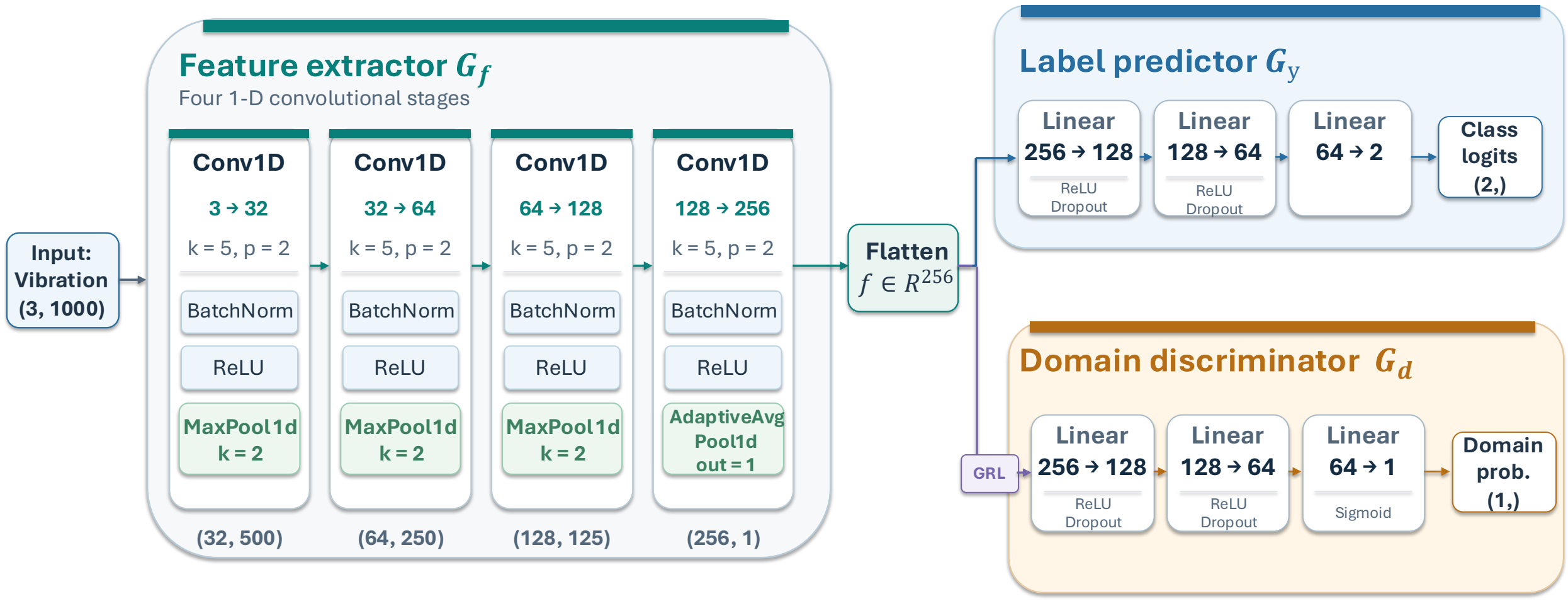

**Fig. 6** Network structure of ADANN

The proposed ADANN was compared with four benchmark approaches. Because each benchmark removes one or more components of Eq. (12), the comparison also serves as an ablation study.

- *Source-Only*. The feature extractor and label predictor are trained exclusively on the labeled source data (Bearing A) and applied directly to the target test set. This benchmark measures the degradation caused by the domain shift between bearings.
- *DANN*. The standard domain-adversarial network of Eq. (8), trained with $\mathcal{L}_y$ and $\mathcal{L}_d$ only. It measures the contribution of adversarial alignment alone.
- *MMD-Only*. Adversarial alignment is replaced by explicit distribution matching: the model is trained with $\mathcal{L}_y$ and $\mathcal{L}_{mmd}$, without the domain classifier.
- *DANN+MMD*. The objective of Eq. (10), trained with $\mathcal{L}_y$, $\mathcal{L}_d$, and $\mathcal{L}_{mmd}$. It differs from ADANN only by the absence of $\mathcal{L}_{ACD}$ and therefore isolates the contribution of the ACD loss.

All benchmarks use the same feature extractor and label predictor architectures and the same training hyperparameters. Each method was trained and evaluated in five independent runs, and we report the mean and standard deviation of precision, recall, and F1-score over the runs on the target test set.

## 4.2 Results

### *4.2.1 Source Labeling*

The results of the unsupervised label identification on Bearing A are shown in Fig. 7. The EWMA chart signals an out-of-control state on all channels, with differences in detection latency and specificity among the channels. The detection delay is defined as the time elapsed between fault occurrence and the first control limit violation. Channel 3 was the most sensitive, triggering an alarm at 32.75 s, a delay of 0.75 s. Channel 2 also detected the fault, with a delay of 3.25 s, so that the average delay over the valid detections was 2 s. Specificity was evaluated through false alarms during the healthy period: Channels 2 and 3 raised no false alarms, whereas Channel 1 triggered three false alarms during the baseline period, the first at 1.25 s, and its alarm was therefore classified as a false positive. These results show that Channel 3 captures the early propagation of this fault type best, and that the flagged segments still require verification. Even so, verifying the flagged segments requires far less effort than labeling the full record manually.

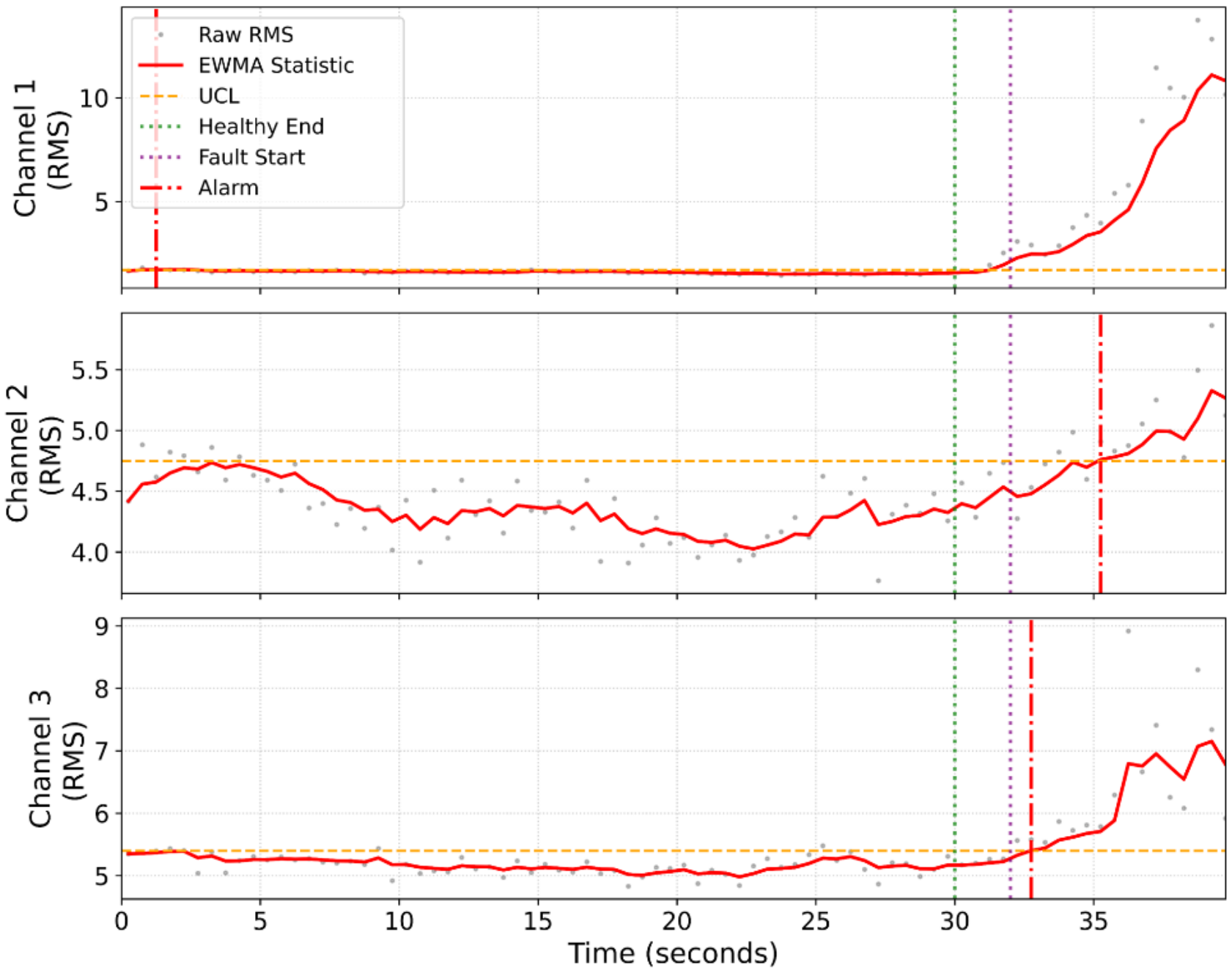

**Fig. 7** EWMA control charts of the RMS series of the three channels of Bearing A (6210 experiment)

### *4.2.2 Early Fault Detection on the Target*

Table 2 reports the precision, recall, and F1-score of all methods on the target test set. All adaptation methods that include the adversarial term improve substantially on the Source-Only baseline, which is highly conservative on Bearing B: it raises almost no false alarms but misses about half of the faulty windows. ADANN's recall and F1-score exceed those of Source-Only and MMD-Only, whereas MMD-Only performs similarly to Source-Only, which indicates that discrepancy minimization alone does not transfer the fault boundary in this case. Under the window labeling of Eq. (2) with the settings of Sect. 4.1, about two thirds of the test windows are faulty (all windows starting from about 18 s onward), so a detector that raises an alarm on every window would attain a precision of 0.67, a recall of 1, and an F1-score of about 0.80. Source-Only and MMD-Only fall below this reference, whereas DANN, DANN+MMD, and ADANN exceed it. Precision separates the methods more clearly: it is 0.67 for the all-alarm detector and 0.9997 for ADANN, which shows that ADANN does separate normal from faulty target windows rather than alarming indiscriminately.

***Table 2*** *Precision, recall, and F1-score on the target test set (mean ± standard deviation over five runs; best mean in bold)*

| **Method** | **Classification Metrics** | | |
|---|---|---|---|
| | Recall | Precision | F1-score |
| Source-Only | 0.4964 ± 0.0448 | **1.0000 ± 0.0000** | 0.6625 ± 0.0390 |
| DANN | 0.7301 ± 0.0476 | 0.9248 ± 0.0783 | 0.8129 ± 0.0273 |
| MMD-Only | 0.5193 ± 0.0434 | 0.9629 ± 0.0823 | 0.6717 ± 0.0276 |
| DANN+MMD | **0.7442 ± 0.0048** | 0.9183 ± 0.0453 | 0.8217 ± 0.0150 |
| **ADANN (proposed)** | 0.7101 ± 0.0243 | 0.9997 ± 0.0006 | **0.8302 ± 0.0166** |

Besides, ADANN attains the highest mean F1-score (0.8302 ± 0.0166). The comparison with DANN+MMD, which differs from ADANN only by the ACD loss, isolates the effect of this term. Adding the ACD loss raises precision from 0.9183 to 0.9997 and lowers recall from 0.7442 to 0.7101. This pattern is the one anticipated by Remark 1: the ACD loss acts on the target's normal data and hence on false alarms, while it provides no target-side information about faults. In short, the ACD loss shifts the operating point of the detector toward fewer false alarms at a modest cost in detection rate, and the F1-score, which weights the two error types equally, is nearly unchanged. Whether this shift is beneficial depends on the relative costs of the two errors, which Sect. 4.3 examines.

### 4.3 Implications for Maintenance Decisions

The detector's output feeds a maintenance decision: each alarm triggers an inspection or intervention, and each missed faulty window is a lost opportunity to act before failure. To make the trade-off in Table 2 concrete, consider an illustrative window-level cost model in which a false alarm costs $c_{FA}$ and a missed faulty window costs $c_{MD}$, with cost ratio $\kappa = c_{FA}/c_{MD}$. For a test set with $N^+$ faulty windows, a detector with precision $P$ and recall $R$ produces $RN^+$ true alarms, $RN^+(1-P)/P$ false alarms, and $(1-R)N^+$ missed windows. Its expected cost per faulty window, in units of $c_{MD}$, is therefore

$$C(\kappa) = \kappa \frac{R\,(1-P)}{P} + (1-R), \qquad (17)$$

which depends only on quantities reported in Table 2. Using the mean values, ADANN has $C(\kappa) = 0.2899 + 0.0002\,\kappa$, compared with $0.2558 + 0.0662\,\kappa$ for DANN+MMD, $0.2699 + 0.0594\,\kappa$ for DANN, $0.4807 + 0.0200\,\kappa$ for MMD-Only, and $0.5036$ for Source-Only. ADANN has the lower expected cost than DANN+MMD whenever $\kappa > 0.52$ and than DANN whenever $\kappa > 0.34$, that is, whenever a false alarm costs more than about half, or one third, of a missed faulty window. It is cheaper than MMD-Only for every $\kappa \geq 0$ and cheaper than Source-Only unless false alarms are about a thousand times cheaper than missed windows.

This analysis is illustrative. It is defined at the window level, whereas maintenance costs in practice depend on the time of the first alarm, on whether repeated alarms during an ongoing fault add cost, and on asset-specific intervention and failure costs; it also uses mean values and does not propagate run-to-run variability. Its purpose is to show that the choice between ADANN and the strongest baselines is a cost-dependent decision rather than a matter of F1-score, and that ADANN is preferable in settings where unnecessary interventions on a healthy asset are costly, for example when an inspection requires a production stop.

### 4.4 Robustness under Synthetic Noise

To evaluate robustness under conditions that are not present in the laboratory data, two extended studies introduce controlled noise into the target data. The first adds Gaussian noise to the normal region and time-varying noise to the fault region (Sect. 4.4.1); the second adds two types of noise commonly encountered in rotating machinery, random shocks and harmonic interference (Sect. 4.4.2). Because the noise is synthetic, these experiments test sensitivity to specific corruptions rather than performance on field data.

### *4.4.1 Gaussian and Time-Varying Noise*

In this study, the roles of the two bearings are reversed so that the uncorrupted Bearing B (6011) serves as the source domain and the noise-corrupted Bearing A (6210) serves as the target domain. The noise is constructed as follows.

- *Normal-region noise.* Zero-mean white Gaussian noise is added to the raw vibration signal in the normal region. Its standard deviation takes four levels, $\sigma_n \in \{2.0, 1.0, 0.5, 0.1\}$, expressed as multiples of the standard deviation of the original signal in the normal region, which simulates varying degrees of sensor noise or environmental interference during data collection.
- *Fault-region noise.* From 32 s onward, a time-varying noise $n(t) = \beta\,(t - t_0)\,\varepsilon(t)$ is added, where $t_0 = 32$ s is the start of the fault region, $\beta \in \{1.0, 0.5, 0.1\}$ controls the growth rate, and $\varepsilon(t)$ is unit white noise. The fault signature thus becomes progressively more pronounced, mimicking degradation processes such as crack propagation or wear accumulation.

The preprocessing, windowing, and labeling procedures of Sect. 4.1 were applied to the corrupted target data, and all methods were retrained and evaluated five times under each condition. Table 3 reports the mean and standard deviation of the F1-scores.

***Table 3*** *F1-scores under Gaussian and time-varying noise (mean ± standard deviation over five runs; best mean in bold)*

| **White Noise Level ($\sigma_n$)** | **Method** | **Scaling Factor ($\beta$)** | | |
|---|---|---|---|---|
| | | $\beta = 1.0$ | $\beta = 0.5$ | $\beta = 0.1$ |
| $\sigma_n = 2.0$ | Source-Only | 0.5717 ± 0.0003 | 0.6352 ± 0.1520 | 0.5714 ± 0.0004 |
| | DANN | 0.5712 ± 0.0009 | **0.7853 ± 0.0235** | 0.5712 ± 0.0006 |
| | MMD-Only | 0.5511 ± 0.0298 | 0.6150 ± 0.1904 | 0.5474 ± 0.0585 |
| | DANN+MMD | 0.5713 ± 0.0009 | 0.7622 ± 0.0304 | 0.5714 ± 0.0003 |
| | **ADANN (proposed)** | **0.5718 ± 0.0175** | 0.7116 ± 0.0541 | **0.5886 ± 0.0547** |
| $\sigma_n = 1.0$ | Source-Only | 0.5711 ± 0.0006 | 0.8247 ± 0.0258 | 0.5387 ± 0.0634 |
| | DANN | 0.5700 ± 0.0018 | 0.8350 ± 0.0092 | 0.5709 ± 0.0016 |
| | MMD-Only | 0.5013 ± 0.0452 | 0.8138 ± 0.0217 | 0.4551 ± 0.2026 |
| | DANN+MMD | 0.5662 ± 0.0093 | 0.8125 ± 0.0478 | 0.5712 ± 0.0008 |
| | **ADANN (proposed)** | **0.5726 ± 0.0041** | **0.8371 ± 0.0254** | **0.6022 ± 0.0976** |
| $\sigma_n = 0.5$ | Source-Only | **0.5717 ± 0.0005** | 0.8530 ± 0.0386 | 0.5560 ± 0.0419 |
| | DANN | 0.5708 ± 0.0014 | 0.8427 ± 0.0116 | 0.5710 ± 0.0013 |
| | MMD-Only | 0.5555 ± 0.0841 | 0.7604 ± 0.1158 | 0.4694 ± 0.1429 |
| | DANN+MMD | 0.5712 ± 0.0051 | 0.8374 ± 0.0112 | 0.5703 ± 0.0033 |
| | **ADANN (proposed)** | 0.5684 ± 0.0623 | **0.8618 ± 0.0363** | **0.6329 ± 0.0845** |
| $\sigma_n = 0.1$ | Source-Only | 0.5461 ± 0.0506 | 0.8378 ± 0.0288 | 0.5494 ± 0.0601 |
| | DANN | 0.5715 ± 0.0002 | 0.8319 ± 0.0139 | 0.5711 ± 0.0007 |
| | MMD-Only | 0.5397 ± 0.0660 | 0.7310 ± 0.0749 | 0.5057 ± 0.0871 |
| | DANN+MMD | 0.5613 ± 0.0478 | 0.8331 ± 0.0101 | 0.5704 ± 0.0021 |
| | **ADANN (proposed)** | **0.6040 ± 0.0766** | **0.8511 ± 0.0301** | **0.6398 ± 0.1257** |

ADANN has the highest mean F1-score in 10 of the 12 settings. For $\beta = 1.0$ and $\beta = 0.1$, most methods obtain F1-scores close to 0.571 with very small standard deviations, which indicates that they converge to the same decision pattern in these settings. ADANN's mean F1-score departs from this value most clearly at $\beta = 0.1$ (0.589 to 0.640), but with large standard deviations, which indicates that ADANN departs from this common pattern in some runs but not consistently.

For $\beta = 0.5$, all adaptation methods separate the classes, and ADANN has the highest mean at $\sigma_n \leq 1.0$ by small margins. At the highest noise level, $\sigma_n = 2.0$, ADANN falls below DANN (0.712 vs 0.785). This result is consistent with Remark 1: the ACD loss anchors the normal region on the target's adaptation data, and when these data are heavily corrupted, the anchor itself is displaced. The quality of the normal target data is therefore a practical precondition for the ACD loss.

#### *4.4.2 Random Shocks and Harmonic Interference*

Two further types of noise are added independently to the target data:

- *Random shocks.* Short-duration, high-amplitude impulses that affect peak values and kurtosis of the vibration signal. The shock occurrence rate is fixed at 100 impacts per second, and the shock amplitude takes three levels, $A_{shock} \in \{3.0, 3.5, 4.0\}$.
- *Harmonic interference.* Narrowband periodic components originating from rotating components such as shafts, gears, and motors, which can mask or mimic fault signatures. The fundamental frequencies are set in the mid range (150, 300, 450, and 600 Hz), and the harmonic amplitude takes three levels, $A_{harmonic} \in \{0.5, 0.8, 1.0\}$.

The procedures of Sect. 4.1 were applied to the corrupted target data. Table 4 reports the mean and standard deviation of precision, recall, and F1-score over five runs for each setting.

***Table 4*** *Precision, recall, and F1-score under random shocks and harmonic interference (mean ± standard deviation over five runs; best mean in bold, ties included)*

| **Parameter** **($A_{shock}/A_{harmonic}$)** | **Method** | **Metrics** | | |
|---|---|---|---|---|
| | | Precision | Recall | F1-score |
| $A_{shock} = 3.0$ | Source-Only | 0.2906 ± 0.0974 | 0.5385 ± 0.3308 | 0.3091 ± 0.1602 |
| | DANN | 0.1862 ± 0.1222 | 0.5770 ± 0.4083 | 0.2783 ± 0.1883 |
| | MMD-Only | 0.4126 ± 0.0629 | 0.7370 ± 0.1514 | 0.5197 ± 0.0559 |
| | DANN+MMD | 0.4217 ± 0.0656 | **0.8593 ± 0.0888** | 0.5637 ± 0.0690 |
| | **ADANN (proposed)** | **0.4562 ± 0.1185** | 0.8505 ± 0.0991 | **0.5758 ± 0.0573** |
| $A_{shock} = 3.5$ | Source-Only | 0.2881 ± 0.0628 | 0.6350 ± 0.0828 | 0.3934 ± 0.0726 |
| | DANN | 0.1898 ± 0.0920 | 0.6485 ± 0.3806 | 0.2924 ± 0.1500 |
| | MMD-Only | 0.3872 ± 0.0539 | 0.8020 ± 0.2151 | 0.5064 ± 0.0531 |
| | DANN+MMD | 0.3820 ± 0.0452 | 0.8488 ± 0.1851 | 0.5237 ± 0.0786 |
| | **ADANN (proposed)** | **0.4086 ± 0.0578** | **0.8528 ± 0.0960** | **0.5480 ± 0.0563** |
| $A_{shock} = 4.0$ | Source-Only | **0.4242 ± 0.1911** | 0.8433 ± 0.1116 | 0.5254 ± 0.1333 |
| | DANN | 0.2013 ± 0.0823 | 0.7020 ± 0.3483 | 0.3118 ± 0.1351 |
| | MMD-Only | 0.3762 ± 0.0280 | 0.9325 ± 0.0712 | 0.5340 ± 0.0198 |
| | DANN+MMD | 0.3505 ± 0.0595 | **0.9713 ± 0.0399** | 0.5110 ± 0.0616 |
| | **ADANN (proposed)** | 0.4039 ± 0.0336 | 0.9262 ± 0.1180 | **0.5582 ± 0.0340** |
| $A_{harmonic} = 0.5$ | Source-Only | 0.2980 ± 0.0278 | **1.0000 ± 0.0000** | 0.4584 ± 0.0323 |
| | DANN | 0.3039 ± 0.0357 | 1.0000 ± 0.0000 | 0.4650 ± 0.0414 |
| | MMD-Only | 0.2722 ± 0.0063 | 1.0000 ± 0.0000 | 0.4278 ± 0.0077 |
| | DANN+MMD | 0.2748 ± 0.0069 | 1.0000 ± 0.0000 | 0.4311 ± 0.0085 |
| | **ADANN (proposed)** | **0.4757 ± 0.1351** | 0.9887 ± 0.0154 | **0.6322 ± 0.1126** |
| $A_{harmonic} = 0.8$ | Source-Only | 0.2667 ± 0.0040 | **1.0000 ± 0.0000** | 0.4211 ± 0.0050 |
| | DANN | 0.2653 ± 0.0028 | 1.0000 ± 0.0000 | 0.4193 ± 0.0035 |
| | MMD-Only | 0.2635 ± 0.0001 | 1.0000 ± 0.0000 | 0.4171 ± 0.0001 |
| | DANN+MMD | 0.2634 ± 0.0000 | 1.0000 ± 0.0000 | 0.4170 ± 0.0000 |
| | **ADANN (proposed)** | **0.3736 ± 0.0594** | 0.9637 ± 0.0561 | **0.5358 ± 0.0663** |
| $A_{harmonic} = 1.0$ | Source-Only | 0.2641 ± 0.0014 | **1.0000 ± 0.0000** | 0.4179 ± 0.0018 |
| | DANN | 0.2665 ± 0.0051 | 1.0000 ± 0.0000 | 0.4209 ± 0.0064 |
| | MMD-Only | 0.2655 ± 0.0018 | 0.9965 ± 0.0064 | 0.4193 ± 0.0020 |
| | DANN+MMD | 0.2651 ± 0.0022 | 0.9988 ± 0.0025 | 0.4189 ± 0.0025 |
| | **ADANN (proposed)** | **0.3999 ± 0.1237** | 0.9155 ± 0.1334 | **0.5390 ± 0.1089** |

Under harmonic interference, all four benchmarks reach a recall of (nearly) 1 with a precision between 0.263 and 0.304. A detector that flags every window has recall 1 and precision equal to the fault prevalence, and the benchmark results are at or close to this reference; for example, DANN+MMD at $A_{harmonic} = 0.8$ has precision 0.2634 with zero variance across runs. The benchmarks therefore raise alarms on nearly all target windows, including the normal ones.

ADANN is the only method that departs from this solution: it raises precision to between 0.374 and 0.476 at a small cost in recall, and its mean F1-score is the highest at all three amplitudes (0.632, 0.536, and 0.539, compared with at most 0.465 for the benchmarks). A plausible explanation, consistent with Remark 1, is that harmonic interference shifts the target's normal features toward the region that the source-trained classifier associates with faults. The benchmarks have no mechanism that evaluates the classifier on target normal data and therefore flag these windows, whereas the ACD loss penalizes exactly these false alarms.

In summary, the robustness experiments support a narrower conclusion than a uniform advantage of ADANN. Where the ACD loss matters, it matters through false alarms: ADANN avoids the all-alarm failure mode under harmonic interference and occasionally under weak time-varying faults, and its effect weakens when the normal adaptation data are themselves heavily corrupted. Under the other synthetic conditions, ADANN performs comparably to the strongest benchmarks.

## 5 Conclusions and Future Work

In this paper, a cross-asset knowledge transfer framework is proposed to achieve robust anomaly detection on a target bearing using only its early-stage normal data. The significance of this framework lies in its practical applicability to real-world industrial scenarios, where labeled fault data for every bearing are unavailable, yet early and accurate fault detection is critical for preventing catastrophic failures and minimizing maintenance costs. By automating the labeling process for the source bearing, the framework alleviates the need for costly and subjective manual annotation. Furthermore, by incorporating both adversarial knowledge transfer and explicit distributional matching through MMD loss, along with a novel adapted classification-domain loss that leverages normal target data, the framework ensures that the diagnostic knowledge learned from the source bearing is effectively transferred and calibrated to the target bearing.

Comprehensive experiments on the publicly available Bearing Cage Full Lifecycle Dataset demonstrated the superiority of the proposed framework. Afterwards, two extended case studies introducing controlled noise into the target domain further validated the framework's robustness. In the future, adaptive mechanisms to dynamically adjust the weight of the ACD loss based on estimated noise levels or the confidence of the normal target data representation could be explored. In addition, the generalizability of the ADANN model to scenarios involving multiple source bearings, significantly different operating conditions, and transfer between fundamentally different components (e.g., from bearings to gears) could be further investigated.

## Declaration of interests

All authors certify that they have no affiliation with or involvement in any organization or entity with any financial interest or non-financial interest in the subject matter or materials discussed in this manuscript.

## Acknowledgements

The research contributions of Yuxuan Li were supported by the Shanghai Pujiang Talents Program (Grant No. 24PJC016) and Youth Science Fund Project (Category C) of the National Natural Science (Grant No. 72501106). The other co-authors did not receive personal financial support from these grants.

## Appendix A Proof of Theorem 1

Throughout the appendix, ln denotes the natural logarithm and $H(p) = -p\ln p - (1-p)\ln(1-p)$ the binary entropy in nats. Write $f = G_f(x)$ for the feature produced from an input window $x$. By the labeling scheme of Sect. 3.3, each source window carries the label normal or abnormal; $p_s^N$ and $p_s^A$ denote the class-conditional densities of the induced features, with proportions $\pi_N$ and $\pi_A = 1 - \pi_N$ fixed by the labels, so that the source marginal is given by Eq. (13). The adaptation set $\mathcal{D}_B^T$ contains normal windows only, so the density $p_t$ of its features is purely normal. Let $D(f) := G_d(f) \in (0,1)$ denote the probability that the domain classifier assigns to the source domain. At the population level, the domain classification loss of Eq. (7) is

$$\mathcal{L}_d = -\mathbb{E}_{f\sim p_s}[\ln D(f)] - \mathbb{E}_{f\sim p_t}\big[\ln\big(1 - D(f)\big)\big]. \qquad \text{(A.1)}$$

In Eq. (12), only the term $-\alpha\mathcal{L}_d$ involves $\theta_d$, so the inner maximization over $\theta_d$ is equivalent to minimizing $\mathcal{L}_d$. The remaining terms do not depend on $\theta_d$: $\mathcal{L}_y$ and $\mathcal{L}_{ACD}$ depend on $(\theta_f, \theta_y)$, and $\mathcal{L}_{mmd}$, the squared MMD of Eq. (9) computed between the marginals $p_s$ and $p_t$, depends on $\theta_f$ only.

Three assumptions are used, corresponding to (i) to (iii) of Theorem 1.

**(i) Capacity.** $G_f$, $G_y$, and $G_d$ are sufficiently flexible that $G_d$ can realize the optimal classifier of Lemma 1, and $G_f$ can realize feature maps under which the densities above exist and the aligned state of Theorem 1 is attainable.

**(ii) Optimal adversary.** At each training step, $G_d$ is updated to the minimizer of $\mathcal{L}_d$ for the current $G_f$ before $G_f$ and $G_y$ are updated. It therefore suffices to analyze the value function $C(\theta_f, \theta_y) := \max_{\theta_d} \mathcal{E}(\theta_f, \theta_y, \theta_d)$.

**(iii) Classification correctness.** At the state of interest, $G_y$ labels all source windows and all target normal windows correctly: there is a measurable abnormal region $R_A$ of the feature space such that

$$p_s^A(R_A) = 1, \qquad p_s^N(R_A) = 0, \qquad p_t(R_A) = 0. \qquad \text{(A.2)}$$

This is the property that $\mathcal{L}_y$ and $\mathcal{L}_{ACD}$ drive toward. Equation (A.2) states it in exact (zero-error) form so that the characterization below is sharp; the same characterization holds approximately when the classification losses are small. We write $R_N$ for the complement of $R_A$.

**Lemma 1** (optimal domain classifier; cf. Goodfellow et al., 2014). For any fixed $G_f$, and hence fixed marginals $p_s$ and $p_t$ regardless of the internal mixture structure of Eq. (13), the domain classifier that minimizes $\mathcal{L}_d$ is

$$D^*(f) = \frac{p_s(f)}{p_s(f) + p_t(f)} \qquad \text{(A.3)}$$

for every $f$ with $p_s(f) + p_t(f) > 0$, and is arbitrary elsewhere. In particular, $D^*(f) = 1/2$ for almost every $f$ if and only if $p_s = p_t$ almost everywhere.

*Proof.* Writing the expectations in Eq. (A.1) as integrals, $\mathcal{L}_d = -\int p_s(f) \ln D(f)\, df - \int p_t(f) \ln\big(1 - D(f)\big) df$. For each fixed $f$ with $p_s(f) + p_t(f) > 0$, the integrand $h(d) = -p_s(f)\ln d - p_t(f)\ln(1-d)$ is strictly convex in $d \in (0,1)$, and its derivative $h'(d) = -p_s(f)/d + p_t(f)/(1-d)$ vanishes at $d^* = p_s(f)/\big(p_s(f) + p_t(f)\big)$, which is therefore the pointwise minimizer. By assumption (i), $D(f)$ can be chosen independently at each $f$, so the pointwise minimizers form the global minimizer $D^*$. If $p_s = p_t$, Eq. (A.3) gives $D^* = 1/2$; conversely, $D^* \equiv 1/2$ forces $p_s = p_t$. ▫

**Lemma 2** (cf. Goodfellow et al., 2014). Let $m = (p_s + p_t)/2$ and let $\text{JSD}(p_s \| p_t) := \frac{1}{2}\text{KL}(p_s \| m) + \frac{1}{2}\text{KL}(p_t \| m) \geq 0$ denote the Jensen-Shannon divergence. Then

$$\mathcal{L}_d(D^*) = \ln 4 - 2\,\text{JSD}(p_s \| p_t) \leq \ln 4, \qquad \text{(A.4)}$$

and consequently the alignment terms of Eq. (12) satisfy

$$-\alpha\mathcal{L}_d(D^*) + \mathcal{L}_{mmd} = -\alpha \ln 4 + 2\alpha\,\text{JSD}(p_s \| p_t) + \mathcal{L}_{mmd} \geq -\alpha \ln 4, \qquad \text{(A.5)}$$

with equality if and only if $p_s = p_t$ almost everywhere.

*Proof.* Since $p_s/(p_s + p_t) = p_s/(2m)$ and $p_t/(p_s + p_t) = p_t/(2m)$, substituting Eq. (A.3) into Eq. (A.1) yields $-\mathbb{E}_{p_s}[\ln D^*(f)] = \ln 2 - \text{KL}(p_s \| m)$ and $-\mathbb{E}_{p_t}\left[\ln\left(1 - D^*(f)\right)\right] = \ln 2 - \text{KL}(p_t \| m)$. Adding the two identities gives $\mathcal{L}_d(D^*) = \ln 4 - 2\,\text{JSD}(p_s \| p_t)$, and the inequality in Eq. (A.4) follows from the non-negativity of the Jensen-Shannon divergence, with equality if and only if $p_s = p_t$. Since $\mathcal{L}_{mmd}$ does not involve $\theta_d$, it passes through the inner maximization unchanged; multiplying Eq. (A.4) by $-\alpha$ and adding $\mathcal{L}_{mmd}$ gives Eq. (A.5). The squared MMD is non-negative and, for the characteristic RBF kernel, vanishes if and only if $p_s = p_t$. Hence both non-negative terms on the right-hand side of Eq. (A.5) vanish exactly when $p_s = p_t$. ▫

**Lemma 3.** Under assumption (iii), the Jensen-Shannon divergence between the two marginals satisfies

$$\text{JSD}(p_s \| p_t) \geq \delta(\pi_A) = \ln 2 - \frac{2 - \pi_A}{2} H\left(\frac{1 - \pi_A}{2 - \pi_A}\right). \qquad \text{(A.6)}$$

The function $\delta$ is continuous and strictly increasing on $[0,1]$, with $\delta(0) = 0$ and $\delta(1) = \ln 2$, and $\delta(\pi_A) = H(\pi_A/2) - \frac{1}{2}H(\pi_A)$, the Jensen-Shannon divergence between $\text{Bernoulli}(\pi_A)$ and $\text{Bernoulli}(0)$. Equality in Eq. (A.6) holds if and only if

$$p_s^N(f) = p_t(f) \quad \text{for almost every } f, \qquad \text{(A.7)}$$

that is, if and only if the source normal features coincide with the target normal features, while the abnormal features occupy the region $R_A$, which is excluded from the target support by Eq. (A.2).

*Proof.* Let $Z$ be a uniform Bernoulli variable indicating the domain, $P(Z=0)=P(Z=1)=1/2$, and let $F$ follow $p_s$ given $Z=0$ and $p_t$ given $Z=1$. The domain posterior $P(Z=0 \mid F=f)=p_s(f)/\big(p_s(f)+p_t(f)\big)$ is the optimal classifier of Eq. (A.3), and the Jensen-Shannon divergence equals the mutual information between domain and feature, $\mathrm{JSD}(p_s \| p_t)=I(Z;F)=\ln 2-H(Z \mid F)$. Now let $V:=\mathbb{I}(F \in R_A)$ indicate that a feature falls in the abnormal region. By the data-processing inequality, $I(Z;F) \geq I(Z;V)=\ln 2-H(Z \mid V)$. Under Eq. (A.2), $P(V=1 \mid Z=0)=p_s(R_A)=\pi_A$ and $P(V=1 \mid Z=1)=p_t(R_A)=0$. Hence $P(V=1)=\pi_A/2$ and $P(V=0)=(2-\pi_A)/2$, which gives $P(Z=0 \mid V=1)=1$, $P(Z=0 \mid V=0)=(1-\pi_A)/(2-\pi_A)$, and

$$H(Z \mid V)=P(V=0)\,H(Z \mid V=0)+P(V=1)\,H(Z \mid V=1)=\frac{2-\pi_A}{2}\,H\left(\frac{1-\pi_A}{2-\pi_A}\right).$$

Combining these steps yields Eq. (A.6). Since $I(Z;V)$ is the Jensen-Shannon divergence between the conditional distributions of $V$ given $Z=0$ and $Z=1$, namely $\mathrm{Bernoulli}(\pi_A)$ and $\mathrm{Bernoulli}(0)$, it also equals $H(\pi_A/2)-\frac{1}{2}H(\pi_A)-\frac{1}{2}H(0)=H(\pi_A/2)-\frac{1}{2}H(\pi_A)$. For monotonicity, differentiate $\delta(\pi_A)=H(\pi_A/2)-\frac{1}{2}H(\pi_A)$ using $H'(p)=\ln\big((1-p)/p\big)$:

$$\frac{d\delta}{d\pi_A}=\frac{1}{2}H'\left(\frac{\pi_A}{2}\right)-\frac{1}{2}H'(\pi_A)=\frac{1}{2}\left[\ln\frac{2-\pi_A}{\pi_A}-\ln\frac{1-\pi_A}{\pi_A}\right]=\frac{1}{2}\ln\frac{2-\pi_A}{1-\pi_A}>0 \quad \text{for } \pi_A \in (0,1),$$

since $(2-\pi_A)/(1-\pi_A)>1$. Because $\delta$ is continuous on $[0,1]$, it is therefore strictly increasing on the whole closed interval. The endpoint values follow from $H(1/2)=\ln 2$ and $H(0)=0$.

For the equality condition, equality in the data-processing step requires the domain posterior to be a function of $V$, that is, constant on $R_A$ and on $R_N$. On $R_A$ it equals 1 because $p_t(R_A)=0$. On $R_N$, where $p_s=\pi_N p_s^N$ by Eq. (A.2), the posterior $\pi_N p_s^N(f)/\big(\pi_N p_s^N(f)+p_t(f)\big)$ is constant in $f$ if and only if $p_t$ is proportional to $p_s^N$ on $R_N$. Since both densities integrate to one over $R_N$ by Eq. (A.2), the proportionality constant is one, which gives Eq. (A.7) and the posterior value $\pi_N/(1+\pi_N)$ on $R_N$. Conversely, under Eq. (A.7) the posterior equals $\pi_N/(1+\pi_N)$ on $R_N$ and 1 on $R_A$, so $H(Z \mid F)=H(Z \mid V)$ and equality holds throughout. ▫

*Proof of Theorem 1.* By assumption (ii) and Lemma 2, the value function of Eq. (12) decomposes as

$$C(\theta_f, \theta_y) = \mathcal{L}_y + \mathcal{L}_{ACD} + \mathcal{L}_{mmd} - \alpha \ln 4 + 2\alpha\, \mathrm{JSD}(p_s \| p_t). \tag{A.8}$$

*Lower bound.* Under assumption (iii), Lemma 3 gives $\mathrm{JSD}(p_s \| p_t) \geq \delta(\pi_A)$, so the domain-adversarial component satisfies $-\alpha\mathcal{L}_d(D^*) = -\alpha\ln 4 + 2\alpha\, \mathrm{JSD}(p_s \| p_t) \geq -\alpha\ln 4 + 2\alpha\, \delta(\pi_A)$, which is Eq. (14); the properties of $\delta$ are those of Lemma 3. In particular, as $\pi_A \to 0$ the gap vanishes and the bound reduces to $-\alpha\ln 4$; by Lemma 2 the unconstrained optimum of the adversarial game is then $p_s = p_t$, with $D^* \equiv 1/2$ by Lemma 1, the classical domain-adversarial equilibrium.

*Attainment.* Suppose that the aligned state of Eq. (A.7) holds: $p_s^N = p_t$ almost everywhere, with the abnormal features confined to $R_A$. Then equality holds in Lemma 3, and the domain-adversarial component equals $-\alpha\ln 4 + 2\alpha\, \delta(\pi_A)$. By Lemma 1, the optimal domain classifier is $D^*(f) = 1$ on $R_A$, where $p_t = 0$, and $D^*(f) = \pi_N/(1 + \pi_N)$ on $R_N$, where $p_s = \pi_N p_t$. Moreover, writing $\mu_s = \pi_N \mu_s^N + \pi_A \mu_s^A$ for the kernel mean embedding of the source marginal and using $\mu_t = \mu_s^N$, which follows from $p_t = p_s^N$, the population MMD evaluates to

$$\mathcal{L}_{mmd} = \|\mu_s - \mu_t\|_{\mathcal{H}}^2 = \pi_A^2 \|\mu_s^A - \mu_s^N\|_{\mathcal{H}}^2 = \mathcal{L}_{mmd}^{\dagger}.$$

Conversely, if Eq. (A.2) holds but $p_s^N \neq p_t$ on a set of positive measure, the equality condition of Lemma 3 fails and the bound in Eq. (14) is strict. Hence, within the class of correctly classifying feature maps, which is the class relevant at deployment, the bound in Eq. (14) is attained exactly at the aligned state $p_s^N = p_t$. ▫